%% file: main.tex
\documentclass[runningheads]{llncs}

\usepackage[year=2024]{eccv}
\usepackage{epsfig}
\usepackage{amsmath, amssymb, booktabs, xcolor, lipsum, graphicx}
\usepackage{arydshln}
\usepackage{array, multirow}
\usepackage{mathtools}
\usepackage{paralist}
\usepackage{enumitem}
\usepackage[normalem]{ulem}
\usepackage{pifont}

\usepackage{array, boldline, makecell, booktabs}
\newcommand{\mycomment}[1]{}
\usepackage{listings}
\usepackage{algorithm}
\usepackage{etoolbox}
\usepackage{multirow}
\usepackage{graphicx}
\usepackage{wrapfig}

\DeclarePairedDelimiter\floor{\lfloor}{\rfloor}
\makeatletter
\AfterEndEnvironment{algorithm}{\let\@algcomment\relax}
\AtEndEnvironment{algorithm}{\kern2pt\hrule\relax\vskip3pt\@algcomment}
\let\@algcomment\relax
\newcommand\algcomment[1]{\def\@algcomment{\footnotesize#1}}
\renewcommand\fs@ruled{\def\@fs@cfont{\bfseries}\let\@fs@capt\floatc@ruled
  \def\@fs@pre{\hrule height.8pt depth0pt \kern2pt}%
  \def\@fs@post{}%
  \def\@fs@mid{\kern2pt\hrule\kern2pt}%
  \let\@fs@iftopcapt\iftrue}
\makeatother

\makeatletter
\DeclareRobustCommand\onedot{\futurelet\@let@token\@onedot}
\def\@onedot{\ifx\@let@token.\else.\null\fi\xspace}

\def\eg{\emph{e.g}\onedot}

\makeatother

\usepackage[pagebackref=true,breaklinks=true,letterpaper=true,colorlinks,bookmarks=false]{hyperref}
\usepackage{eccvabbrv}

\input{preamble}
\usepackage[accsupp]{axessibility}  % Improves PDF readability for those with disabilities.
\usepackage{}

\usepackage{orcidlink}

\begin{document}

% ---------------------------------------------------------------
\title{ActionSwitch: Class-agnostic Detection \\ of Simultaneous Actions in Streaming Videos} 
% \title{STOV-TAL: Towards Open-Vocabulary Temporal Action Localization with Self-Training}

\titlerunning{ActionSwitch}

\author{Hyolim Kang\orcidlink{0000-0003-1571-4359} \and
Jeongseok Hyun\orcidlink{0000-0002-8629-3929} \and
Joungbin An\orcidlink{0000-0003-0418-900X} \and \\
Youngjae Yu\orcidlink{0000-0002-5867-0782} \and
Seon Joo Kim\orcidlink{0000-0001-8512-216X}} 

\authorrunning{Kang. et al.}

\institute{Yonsei University}

\maketitle

\input{sec/0_abstract}
\input{sec/1_intro}

\input{sec/2_rel_works}
\input{sec/3_methodology}
\input{sec/4_experiments}

\input{sec/5_conclusion}

\noindent\textbf{Acknowledgement}
This work was supported by Institute of Information \& communications Technology Planning \& evaluation (IITP) grant funded by the Korea government (MSIT) (No.RS-2022-II220113, Developing a Sustainable Collaborative Multi-modal Lifelong Learning Framework), the National Research Foundation of Korea (NRF) grant funded by the Korea government (MSIT) (NRF- 2022R1A2C2004509), and Artificial Intelligence Graduate School Program,
Yonsei University, under Grant 2020-0-01361

\bibliographystyle{splncs04}
\bibliography{main}
\end{document}

%% file: preamble.tex
\usepackage{arydshln} % for cdashline - this package should be loaded before [xcolor,colortbl]. Otherwise, conflict with multirow

\usepackage[dvipsnames]{xcolor, colortbl}

\usepackage{epsfig}
\usepackage{graphicx}
\usepackage{amsmath}
\usepackage{amssymb}
\usepackage{dashrule}
\usepackage{booktabs}
\usepackage{array, multirow}
\usepackage{mathtools}
\usepackage{enumitem}
\usepackage{float}
\usepackage[normalem]{ulem}
\useunder{\uline}{\ul}{}
\graphicspath{ {./pics/} }
\usepackage{bbm}

\usepackage{pifont}

\usepackage{tikz}
\usetikzlibrary{calc}
\usetikzlibrary{shapes}
\usepackage{lipsum}

\definecolor{brightlavender}{rgb}{0.75, 0.58, 0.89}

%% file: sec/0_abstract.tex
\begin{abstract}

Online Temporal Action Localization (On-TAL) is a critical task that aims to instantaneously identify action instances in untrimmed streaming videos as soon as an action concludes---a major leap from frame-based Online Action Detection (OAD).
Yet, the challenge of detecting overlapping actions is often overlooked even though it is a common scenario in streaming videos.
Current methods that can address concurrent actions depend heavily on class information, limiting their flexibility.
This paper introduces ActionSwitch, the first class-agnostic On-TAL framework capable of detecting overlapping actions.
By obviating the reliance on class information, ActionSwitch provides wider applicability to various situations, including overlapping actions of the same class or scenarios where class information is unavailable.
This approach is complemented by the proposed ``conservativeness loss'', which directly embeds a conservative decision-making principle into the loss function for On-TAL.
Our ActionSwitch achieves state-of-the-art performance in complex datasets, including Epic-Kitchens 100 targeting the challenging egocentric view and FineAction consisting of fine-grained  actions. 
\keywords{Online Video Understanding \and Class-agnostic Detection}
\end{abstract}

%% file: sec/1_intro.tex
\section{Introduction}
\label{sec:intro}

\begin{figure}[t]
    \includegraphics[width=\textwidth]{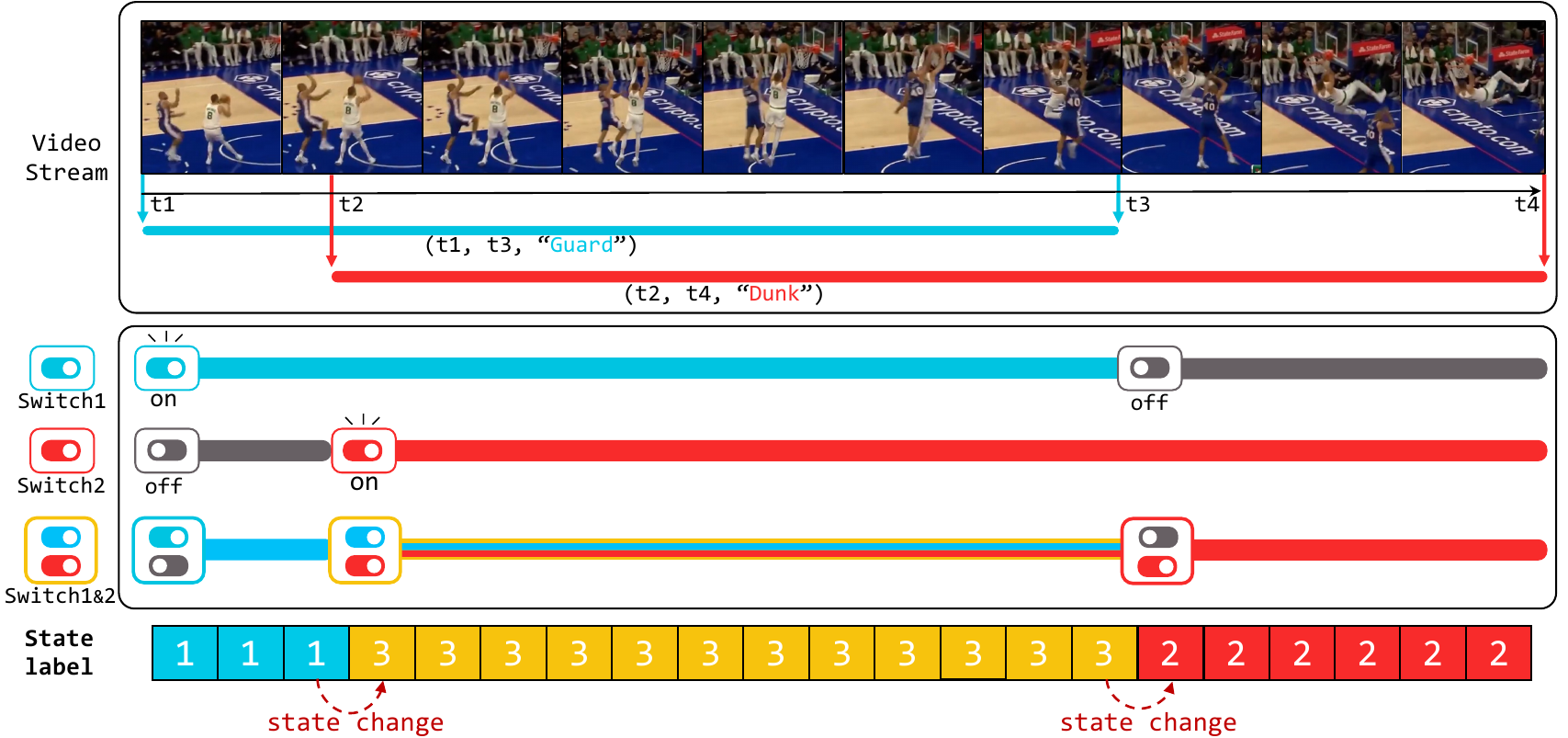}
    \caption{
    Overview of the ActionSwitch Framework: State label is derived from the \textit{sum of the ids of activated switches}. For example, the state is labeled as `3' between t2 and t3 when switches `1' and `2' are simultaneously active, whereas it registers as `2' from t3 to t4 when only switch `2' is active. 
    State changes signify action instance boundaries, and our `conservativeness loss' minimizes state fluctuations to improve detection accuracy.
    }
    \label{fig:teaser}
\end{figure}

As the demand for autonomous driving systems \cite{autonomousdriving}, robotics applications \cite{lee_robot}, and ego-centric perception \cite{epickitchen} continues to grow, the importance of online video understanding tasks has become more pronounced. 
Online Action Detection (OAD) \cite{oad} is a pivotal online video understanding task that has gained considerable attention \cite{red, trn, oadtr, lstr, ts-oad, woad, miniroad}, as it aims to detect the action class of each frame from a streaming video. 
However, merely assigning class scores to individual frames does not capture the essence of actions in the video, which inherently extend over multiple frames and form \textit{action instances}.
In particular, this instance-level understanding is a de facto standard in conventional video understanding tasks, including Temporal Action Localization (TAL) \cite{actionformer, tadtr, gtad, bmn, bsn++}.

To provide instance-level understanding in streaming videos, Online Temporal Action Localization (On-TAL) \cite{cagqil, simon, oat, 2pesnet} has recently been proposed. 
Its goal is to instantaneously identify action instances in untrimmed streaming video, accurately determining the start and end times, and classifying each instance as soon as it concludes.

While the use of (soft-) non-maximum suppression (NMS)~\cite{softnms} is indispensable in conventional TAL, the online constraints of On-TAL preclude any retrospective modifications after the initial generation of action instances. 
This requisite, coupled with the need for instant detection, poses challenges in directly applying TAL methods to On-TAL.

Thus, a straightforward approach is extending the OAD framework; this involves thresholding the per-frame output from the OAD model and aggregating these outputs to generate action instances.
Nonetheless, a simple threshold-based grouping of either a class-agnostic or class-aware OAD model gives rise to two significant challenges: i) the difficulty in detecting actions occurring simultaneously and ii) the generation of fragmented and noisy action instances \cite{cagqil}. 
The main objective of this work is to explore these challenges and offer effective solutions.

% Detecting overlapping action instances in streaming videos remains relatively underexplored, even though considering only a single action at a given time contradicts the intuition that actions are intricately connected~\cite{multithumos}.
Detecting overlapping action instances in streaming videos, despite their frequent occurrence, remains relatively underexplored. This aspect holds considerable importance as it harmonizes with the intuition that actions are intricately connected~\cite{multithumos}.
In particular, an extension of a class-agnostic OAD model \cite{cagqil} is entirely incapable of recognizing concurrent actions.
While a class-aware model \cite{simon} may attempt to do so, it faces considerable limitations.
This is largely because class-aware models rely on independently grouping per-class scores to generate action instances.
As a result, their ability to detect overlapping instances is intricately tied to the predefined action classes in a dataset.
The Epic-Kitchens dataset \cite{epickitchen}, with its many overlapping instances within the same class, highlights this limitation that class-aware OAD models struggle to address.
Additionally, these models are highly sensitive to predetermined thresholds, a challenge that intensifies with an increase in the number of classes.
Moreover, the dependence on class information to distinguish action instances presents difficulties in general On-TAL applications, especially when predefining all possible action classes is not practical.
The unique advantage of disentangling class information from action proposal generation is further elaborated in Sec.~\ref{sec:class-agnostic-detection}.

To this end, we propose the first class-agnostic On-TAL framework capable of detecting overlapping action instances. We begin by considering a machine with multiple switches (Fig. \ref{fig:teaser}), namely \textbf{ActionSwitch}. Each switch performs class-agnostic yet mutually exclusive action instance detection where the activated switch indicates detecting ongoing action.
By incorporating multiple switches, our system adeptly detects overlapping instances all without necessitating class information. 
We devise a finite state machine corresponding to our ActionSwitch framework to instantiate this concept. 
Leveraging a conventional OAD framework that outputs state labels for each frame, we create what we refer to as a state-emitting OAD.
We can effectively generate action instances online by appropriately grouping the framewise output of the state-emitting OAD model, even if there are overlaps among them.

Another remaining challenge is noisy and fragmented action proposal generation, a byproduct of the discrepancy between the frame-centric nature of OAD and the instance-level demands of On-TAL.
From an instance-level perspective, even a single erroneous decision can shatter the continuity of an action instance, which is particularly problematic when identifying long action instances.
Consequently, the agent must be conservative in altering its decisions.
Past approaches \cite{cagqil, simon} learned this principle from data by modeling decision context, yet we advocate for a more direct method: directly infuse conservatism into the loss function.
We propose a \textbf{Conservativeness loss}, an auxiliary loss term that encourages the agent to depend on its previous decision. 
% This method seamlessly merges the principle of conservatism by simply adding our auxiliary loss to the conventional loss, obviating the need for intricate architectural changes.
This method integrates conservatism into the standard loss function without necessitating complex architectural modifications.

We conduct comprehensive experiments on three primary action localization datasets \cite{THUMOS14, fineaction, epickitchen} to validate our proposed method's efficacy and establish robust baselines for future research of On-TAL. It is also worth noting that while the majority of action localization datasets (including the above three) require the ActionSwitch framework with just two switches, three or more switch configurations are also tested in MultiTHUMOS \cite{multithumos} dataset, providing valuable insight for further research. \\

%Our model's effectiveness has been thoroughly validated through rigorous evaluation with the standard THUMOS'14 dataset, the extensively annotated MultiTHUMOS dataset, and the significantly larger and more intricate FineAction and EpicKitchens datasets, which collectively demonstrate the robustness of our proposed methodology.

\noindent Our contributions can be summarized as follows:
% \begin{center}
\begin{compactitem}
% \begin{itemize}
    \item We introduce \textit{ActionSwitch}, the first class-agnostic On-TAL framework that is capable of detecting overlapping action instances by incorporating a finite state machine concept.
    \item We introduce \textit{Conservativeness loss} to address the challenge of noisy and fragmented action proposals, effectively incorporating conservatism directly into the loss function with minimal modification.
    \item We demonstrate the effectiveness of ActionSwitch through experimental results from multiple action localization datasets \cite{THUMOS14, epickitchen, multithumos, muses, fineaction} and extensive ablation studies.
% \end{itemize}
\end{compactitem}
% \end{center}

%% file: sec/2_rel_works.tex
\section{Related Work}

\subsection{Streaming Video Understanding}
Online Action Detection (OAD) \cite{oad} is a well-established task in online video understanding that requires identifying the action class of the current input frame in a streaming setting. A plethora of work \cite{trn, idn, colar, oadtr, lstr, gatehub, testra, miniroad} has been introduced, mainly focusing on temporal modeling of the past visual context. Despite the progress, the frame-centric approach of OAD and its evaluation with per-frame mean Average Precision (mAP) falls short of addressing the needs of real-world streaming video analysis that necessitates instance-level recognition.

Online Detection of Action Start (ODAS) \cite{odas, startnet}, on the other hand, is a task that pinpoints the initial timestep of each action within a streaming video. By retrieving one unique timestep for each action instance, ODAS implicitly conveys the concept of action instances. However, its concentration solely on the initiation point of action instance restricts its broader applicability.

\subsection{Online Temporal Action Localization}
Online Temporal Action Localization (On-TAL) targets real-time identification of action instances in streaming video.
A direct method involves extending the OAD framework~\cite{cagqil, simon} for On-TAL, which necessitates incorporating the agent's decision history to produce accurate action instances.
Several approaches have been devised to tackle this challenge, including the incorporation of a unique grouping module complemented by a distinct training strategy \cite{cagqil}, as well as the implementation of a decision context token \cite{simon}.
The exact processes for generating action instances in these methods are detailed further in our supplementary materials to
make the paper self-contained.

On the other hand, the TAL extension has been adapted for On-TAL using data-driven online filtering \cite{oat}, which mimics the online version of NMS. Yet, all these methods rely on class-specific information to separate concurrent actions. For example, SimOn \cite{simon} groups the same class decisions to generate action instances whereas OAT \cite{oat} uses a handcrafted threshold value to explicitly suppress overlapping action instances of the same class, indicating that neither can handle overlapping action instances that have the same action class. In contrast, our ActionSwitch framework is the first On-TAL approach to identify overlapping actions irrespective of class information.

\subsection{Class-agnostic Detection}
\label{sec:class-agnostic-detection}
Class-agnostic proposal generation is already a widely accepted standard in both Object Detection (OD) \cite{rfcn, fast-rcnn, efficientdet} and TAL \cite{pgcn, gtad, bmn} literature, and recent works highlight the unique benefits of disentangling the proposal generation and classification. OLN \cite{oln} suggests that focusing on localization without class constraints enhances generalization, and Maaz et al. \cite{classagnostictransformer} show benefits of class-agnostic detector for open-world detection and self-supervised learning \cite{detreg}. In videos, recognizing class-agnostic event boundaries, as discussed in \cite{gebd}, aligns with human perception, which does not depend on predefined action classes. These recent advances \cite{classagnosticobjectdetection, openworldinstanceseg} and flourishing development of video-language models \cite{vificlip} validate our approach to separate action proposal generation from classification in On-TAL, indicating the potential for a more flexible framework suited to open-world and open-vocabulary contexts.

%% file: sec/3_methodology.tex
\section{Methodology}
\label{sec:methodology}
\subsection{Problem Setting}
\label{sec:probsetting}
\noindent\textbf{On-TAL} Let us consider an untrimmed video $V=\{x_{\tau}\}^T_{\tau=1}$ with $M$ action instances $\Psi=\{\psi_m\}_{m=1}^M=\{(t^s_m, t^e_m, c_m)\}_{m=1}^M$ is given in a streaming format.
$x_{\tau}$ indicates $\tau$th frame, $t^s_m$, $t^e_m$ represent the start and the end timestep, and $c_m$  is the class label of the $m$th action instance $\psi_m$.
Following the previous convention \cite{gtad, bmn, cagqil, oadtr}, consecutive $k$ frames are converted to a $D$ dimensional visual feature sequence $f \in \mathbb{R}^{\floor{\frac{T}{k}} \times D}$ with a pretrained snippet encoder.
Subsequent operations, including online action instance generation, are performed on this feature sequence. 
The goal of On-TAL task is to generate and accumulate action proposals $\psi$ as soon as their completion is detected, aiming to reconstruct $\Psi$ \textit{without retrospective modification} to each $\psi$.

%In this paper, we introduce a new condition to the previous formulation to incorporate state-awareness.
%In this setting, an On-TAL model is deemed state-aware if it can identify the start timestep $s_m$ of the action instance $\psi_m=(s_m, e_m, c_m)$ at the current timestep $t$,  where $s_m \leq t < e_m$.
%Having the ability to instantaneously pinpoint the start timestep enables the agent to monitor the status of ongoing actions, including their progression. Notably, the state-aware On-TAL problem formulation fully encompasses the ODAS problem setting as per its definition. 
%Since decoupling the generation of action instances from their classification is one of our desiderata, we focus on performing state-aware and class-agnostic On-TAL.
% Our primary goal is to perform state-aware and class-agnostic On-TAL, since decoupling the generation of action instances from their classification is one of our desiderata.

% 여기 아래 문장들 다 너무 길어요. 모든 문장들이 something, something 이런식으로 중간에 콤마 넣으면서 너무 길게 늘어져있어요. 일단 임시로 세미콜론이나 period 로 좀 나눠놨어요.

\noindent\textbf{OAD-extension vs TAL-extension} While On-TAL requires immediate identification of action endpoints, it allows flexible timing in identifying action startpoints. 
This flexibility is exploited by the TAL-based approach \cite{oat}; it identifies action starts at the time of action instance generation, thereby benefiting from rich context.
In contrast, OAD-based methods \cite{cagqil, simon} are inherently designed to recognize the start of actions in real-time, often working within a highly constrained initial context. 
This distinction emphasizes the unique role of OAD-based methods in On-TAL. They are able to adeptly handle early action detection, a task TAL-based methods cannot perform.
% In particular, the ability to promptly identify action starts can extend the On-TAL model's utility to scenarios like ODAS without any modification, where the urgency of detecting an action's start is important.
In particular, promptly identifying an action's start can extend the On-TAL model's utility to scenarios where immediate detection of an action's initiation is critical, an aspect tackled in Online Detection of Action Start (ODAS) \cite{odas}. 

\subsection{State-emitting OAD Model}
In order to address the class-agnostic On-TAL problem, we begin by extending class-agnostic OAD models, following the baseline approach described in \cite{cagqil}. In this setting, the OAD model produces a framewise actionness decision; 0 indicates the absence of action, and 1 the presence of action. Action start and end can be instantaneously determined by detecting changes from 0 to 1 and 1 to 0 respectively. Additionally, the progress status of an action, which is helpful for many practical applications, can be easily calculated by measuring the distance between the action's start timestep and the current timestep.
\begin{figure}[t]
    \centering
    \includegraphics[width=1\linewidth]{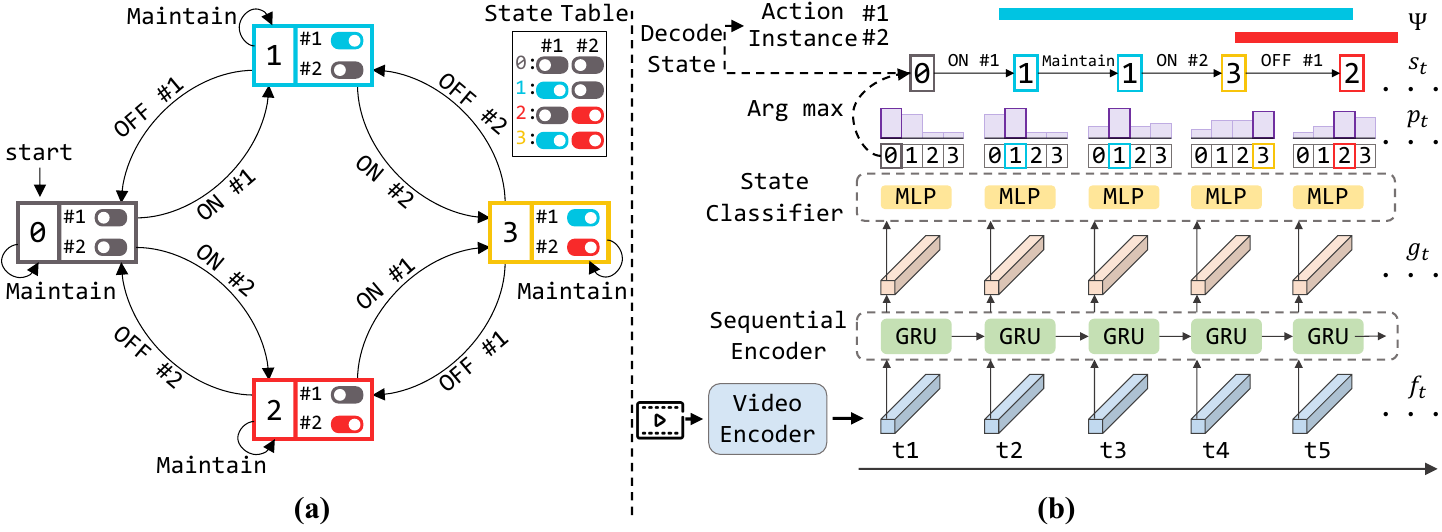} % 최대한 키워본게 이 버전
    \caption{
    (a) State diagram of ActionSwitch framework. Some connections are omitted for simplicity. (b) Overall architecture of state-emitting OAD model.
    }
    \label{fig:main}
\end{figure}

The primary challenge with this approach is its inability to detect overlapping action instances, as decisions are based solely on the presence of the action in the current frame.
Aggregation of these binary action decisions tends to merge multiple overlapping instances into a single instance.
A possible solution is to use multiple OAD models for mutually exclusive action instance detection.
For instance, in a setup with two OAD models and a single action, one model should signal ``no action'' when the other detects the action.
However, this interdependence of the models' decisions makes the implementation complicated.
% For instance, with two OAD models and one ongoing action, if one model detects the action, the other must register 'no action'. 
% However, the practical implementation of this approach is not straightforward due to the interdependency in the models' decisions.
% 위 문장에서 trivial 이라고 한 부분 뭔가 납득이 잘 안되는거 같아요. 좀더 설명하면 좋을듯 hyo:했어

Instead, we abstract the concept into a \textit{single} machine with multiple switches (Fig.~\ref{fig:main} (a)).
A two-switch configuration is exemplified for simplicity, although configurations with three or more switches are also feasible (See Section \ref{sec:ablation_studies}).
Switch 1 is activated when the first action is detected, and Switch 2 is activated when the other action is detected while Switch 1 is still active. 
The finite state machine corresponding to this machine has four states: i) no switch activated, ii) switch 1 activated, iii) switch 2 activated, and iv) both switches activated.
These states are illustrated in the \texttt{State Table} in Fig.~\ref{fig:main} (a). 
State transitions here indicate the commencement or termination of a certain action instance, enabling the model to determine the start, end, or progress status of multiple ongoing actions in real-time by analyzing the stored state history. 
%It is important to note that three state formulation that just tracks the number of switches pressed in a certain timestep is insufficient for accurate action instance detection because of the lack of unique decoding in some overlapping cases.

In order to realize this concept using neural networks, we need to make an important design choice regarding the interpretation of the network's output. For simplicity, we directly interpret the output of the network as the state label, yielding a state-emitting OAD model (Fig.~\ref{fig:main} (b)). At each timestep $t$, $f_t \in \mathbb{R}^D$ is fed into the uni-directional sequential encoder, producing hidden state $g_t \in \mathbb{R}^D$.
Subsequently, the categorical probability distribution $p_t$ is generated by $p_t=\textit{softmax}(\texttt{SC}(g_t))$, where \texttt{SC} denotes a State Classifier $\texttt{SC}:\mathbb{R}^D \to \mathbb{R}^S$ that can be any arbitrary neural network, with $S$ representing the number of states. 
The state corresponding to the current timestep $t$ is obtained by simply calculating $s_t=\texttt{argmax}(p_t)$.
Note that \texttt{argmax} operation here eliminates the need for handcrafted threshold selection, which was the primary problem of previous works \cite{oat, simon}.
\begin{figure}[t]
    \centering
    \includegraphics[width=0.7\linewidth]{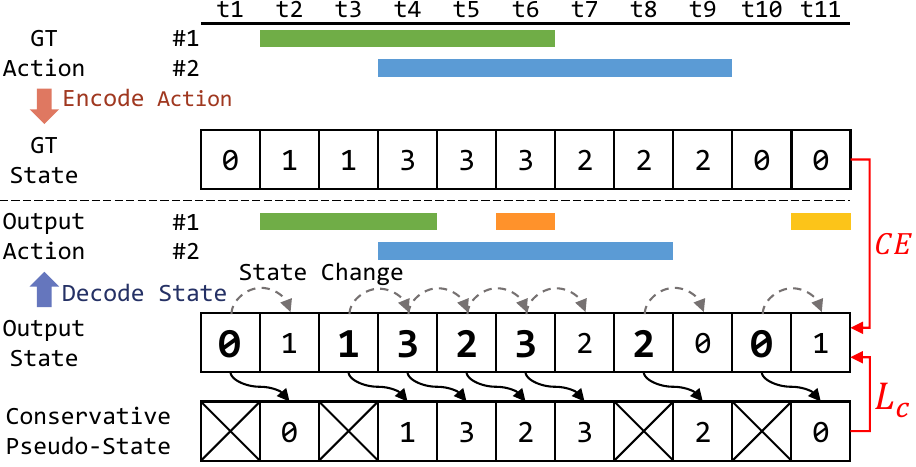}
    \caption{
    %notation... (loss, 오른쪽에)
    Training process in ActionSwitch.
    $CE$ and $\mathcal{L}_c$ denote the terms in Eq.~\ref{eq:loss_function}.
    % \textcolor{blue}{loss notation corresponds to Eq.2, place it on the right side of red arrow}
    GT state and conservative pseudo-state are used for training.
    GT states are encoded from GT action instances while the pseudo-states come from the model's own predictions.
    At \texttt{t6}, action 1 is temporarily lost and results in the fragmentation of the action instance. However, with our conservativeness loss, the output of action instances becomes robust against such fragmentation (\texttt{t6}) and noisy output (\texttt{t11}).
    }
    \label{fig:cons_loss}
\end{figure}

%The first choice is to treat the output as a state transition, where 0 refers to maintaining the current state, while 1 and 2 indicate state transitions made by manipulating switch 1 and 2, respectively. However, since state transitions are rare, this approach results in severe class imbalance, which is not ideal in a supervised learning setting. To address this issue, This approach requires one additional state compared to the former approach, but it greatly reduces the class imbalance problem.

\subsection{Conservativeness Loss}
\label{sec:conservative_loss}
\begin{algorithm}[t!]
\caption{Conservativeness loss in a PyTorch style.}
\label{alg:cons_loss}
\definecolor{codeblue}{rgb}{0.25,0.5,0.5}
\lstset{
  backgroundcolor=\color{white},
  basicstyle=\fontsize{7.5pt}{7.5pt}\ttfamily\selectfont,
  columns=fullflexible,
  breaklines=true,
  captionpos=b,
  commentstyle=\fontsize{7.2pt}{7.2pt}\color{codeblue},
  keywordstyle=\fontsize{7.2pt}{7.2pt},
  texcl=<true|false>
}
\begin{lstlisting}[language=python]
# logits: tensor in the shape of (B, L, n_state)
# return loss penalized by context change (cc)
pred_state = torch.argmax(logits, dim=2) # (B, L)
cc_mask = pred_state[:, 1:] != pred_state[:, :-1]
cc_targets = pred_state[:, :-1][cc_mask]
cc_logits = logits[:, 1:][cc_mask]
return F.cross_entropy(cc_logits, cc_targets)
\end{lstlisting}
\end{algorithm}

Action boundaries are typically much less frequent than non-boundary frames, which is a widely observed phenomenon in most videos. 
It is evident that leveraging this prior is advantageous in grouping framewise OAD decisions in generating action instances. 
Previous methods \cite{cagqil, simon} attempted to learn this prior by modeling decision history. 
Our approach, on the other hand, involves encoding it directly into the loss function, thereby leading to the novel conservativeness loss term $\mathcal{L}_c$:
\begin{equation}
\mathcal{L}_c(p_t, s_{t-1}) =
\begin{cases}
    -log(p_t[s_{t-1}]), & \text{if } \texttt{argmax}(p_t) \neq s_{t-1}; \\
    \makebox[\widthof{$-\log(p_t[s_{t-1}])$}][c]{\phantom{-}0}, & \text{otherwise.}
    % \text{if } \texttt{argmax}(p_t) = s_{t-1}.
\end{cases}
\end{equation}
This loss term penalizes state change by applying standard cross-entropy loss with a pseudo-label $s_{t-1}$, an output from the model's own prediction in the previous step. 
Note that this loss term is only imposed at the context-changing timestep, as shown in Fig.~\ref{fig:cons_loss}.
A significant advantage of this loss term is its simplicity; it integrates seamlessly into the standard OAD framework for On-TAL extension without necessitating any additional modules or architectural modifications.
Furthermore, the loss term can be easily implemented with just five lines of code (Algorithm \ref{alg:cons_loss}), making it a practical and efficient solution.

With a ground truth state label $y_t$, final loss term $\mathcal{L}$ for each timestep $t$ is defined as follows:
\begin{equation}
\label{eq:loss_function}
\mathcal{L}(p_t, y_t, s_{t-1} ) = CE(p_t, y_t) + \alpha \mathcal{L}_c(p_t, s_{t-1}),
\end{equation}
where $CE$ refers to the typical cross entropy term and $\alpha$ denotes a weight to balance both losses. For training, we generate a framewise ground truth state label from the ground truth action instances (\texttt{Encode Action} in Fig. \ref{fig:cons_loss}), assuming that switch 1 is first activated and then switch 2 to avoid ambiguity.
\mycomment{
\begin{algorithm}[t!]
\caption{\texttt{get\_4state\_proposals} in Python.}
\definecolor{codeblue}{rgb}{0.25,0.5,0.5}
\lstset{
  backgroundcolor=\color{white},
  basicstyle=\fontsize{7.5pt}{7.5pt}\ttfamily\selectfont,
  columns=fullflexible,
  breaklines=true,
  captionpos=b,
  commentstyle=\fontsize{7.2pt}{7.2pt}\color{codeblue},
  keywordstyle=\fontsize{7.2pt}{7.2pt},
  texcl=<true|false>
}
\begin{lstlisting}[language=python]
def get_4state_proposals(action_list):
    #IN: action_list (List), ex [0,1,1,3,3,2,2,2,0], contain state labels in {0,1,2,3,}
    #OUT: proposals (List), ex [[1,5],[3,8]]
    #turn off button2 and make button1_proposals...
    convert_dict = {0: 0, 1: 1, 2: 0, 3: 1}
    button1_proposals = get_2state_proposals([convert_dict[i] for i in action_list])
    #turn off button1 and make button2_proposals...
    convert_dict = {0: 0, 1: 0, 2: 1, 3: 1}
    button2_proposals = get_2state_proposals([convert_dict[i] for i in action_list])
    #merge two lists...
    button1_proposals.extend(button2_proposals)
    return button1_proposals
\end{lstlisting}
\label{alg:4state}
\end{algorithm}
}

During the inference stage, we store $s_t$ for each timestep $t$ in a history queue. By comparing $s_t$ and $s_{t-1}$, we can instantly infer the start and end timesteps. The history queue can then be decoded into action instances using a straightforward algorithm (\texttt{Decode State} in Fig.~\ref{fig:main} (b) and \ref{fig:cons_loss}), which we elaborate on in our supplementary material.
This approach ensures ``no room for boundary mismatches'', as the predicted state label sequence corresponds to one clear action scenario, and hence does not need a separate boundary-matching module.
It significantly differentiates our approach from other boundary-matching based algorithms~\cite{bsn, bmn}, which involve exhaustive matching between boundaries and filtering processes.

%For clarity, we provide an explicit algorithm of the decoding process in the 2 switch configuration in Algorithm~\ref{alg:4state}.
%First, the function \texttt{get\_2state\_proposals} finds all continuous \texttt{'1'} sequences.
%To clarify with an example, given the input list \texttt{[0,1,1,0,0,1,1,1,0]}, the output of \texttt{get\_2state\_proposals} will be \texttt{[[1,3],[5,8]]} because there is a sequence of \texttt{1}s from index 1 to 3 and another from 5 to 8.
%Then, the \texttt{get\_4state\_proposals} function applies this function to a dual-switch setup, deactivating one switch at a time and then integrating the results.
%This approach ensures no room for boundary mismatches, as the predicted state label sequence corresponds to one clear action scenario, and hence does not need a separate boundary-matching module.

%% file: sec/4_experiments.tex
\section{Experiments}
In this section, we present the experimental results and ablation studies of our proposed method, mainly focusing on the On-TAL performance.
Moreover, we include results from the ODAS benchmark for a comprehensive evaluation. 
%In this section, we present the experimental results and ablation studies of our proposed method, mainly focusing on the On-TAL performance. For each dataset, we carefully tune prior works \cite{simon, oat, cagqil} to achieve the optimal F1 scores, ensuring a fair comparison. Moreover, since our method extends from OAD frameworks, it is inherently equipped to address the ODAS task. Accordingly, we include results from the ODAS benchmark for a comprehensive evaluation. 

\subsection{Datasets and Features}
\noindent\textbf{Datasets} To validate the ActionSwitch framework's efficacy and general applicability, we evaluate our method on the variety of popular TAL datasets, including standard THUMOS14 \cite{THUMOS14}, large-scale FineAction \cite{fineaction}, and large egocentric Epic-Kitchens 100 \cite{epickitchen}.
Additional experimental results on the MUSES \cite{muses} dataset and a detailed analysis of each dataset in terms of appropriately evaluating On-TAL are provided in the supplementary materials.

% Our proposed method's performance is assessed on two public datasets, THUMOS14~\cite{thumos14} and FineAction~\cite{fineaction}.
% Further elaboration on the datasets, our rationale for selecting them, and additional experimental results obtained from the MUSES~\cite{muses} dataset are included in the supplementary material.

\noindent\textbf{Features} 
For the THUMOS14 dataset, we adopt the two-stream TSN model~\cite{tsn} trained on Kinetics~\cite{i3d} to extract features, following~\cite{cagqil, simon, gtad, oat}.
For the MUSES dataset, we use the officially available features for our experiments. 
However, in the FineAction dataset, the provided features are too coarse-grained (16 frames per one feature vector), which results in an insuperable bottleneck for detecting fine-grained action instances. To address this issue, we follow ~\cite{miniroad} and increase the temporal resolution of the features by a factor of four.
For the Epic-Kitchens 100 dataset, we used publicly available\footnote{https://pytorchvideo.org/} Kinetics400 pretrained Slowfast \cite{slowfast} network weight with stride four to extract features. It is important to note that all experiments in the same table are conducted with the \textit{same features}, whether for offline or online TAL methods to ensure a fair comparison.

\subsection{Evaluation Metric}
\noindent\textbf{F1 score for class-agnostic evaluation} 
In various TAL literatures \cite{bsn, bsn++, bmn, rtdnet, butal}, Average Recall (AR) under varying Intersection over Union (IoU) thresholds is used to evaluate class-agnostic action proposal generation.
This metric is suited to standard TAL, where the output from proposal generators undergoes NMS-like processing to eliminate redundant detection. 
However in the On-TAL context, as discussed in Section \ref{sec:probsetting}, such retrospective modifications to generated proposals are not allowed, making it essential to assess both the precision and recall.

Therefore, we use the \textbf{F1 score} as the main evaluation metric in the class-agnostic On-TAL to capture the balance between precision and recall. 
We first run the Hungarian algorithm \cite{hungarian} to provide optimal bipartite matching between the ground truth action instances and predictions based on their temporal IoU (tIoU).
Then, a prediction is considered a true positive based on whether it surpasses a certain tIoU threshold.
Note that using Hungarian matching here is in the same context as using it in popular query-based object detection methods \cite{detr}. For the sake of concise demonstration, F1 and recall are mainly presented at a setting of tIOU=$0.5$. F1 and recall in other tIOU thresholds, straightforward pseudocode of the metric calculation, and interesting discussions about the F1 score and its relationship with standard classwise mAP metric can be found in the supplementary materials. 

\noindent\textbf{mAP metric for class-dependent evaluation} 
To provide an apple-to-apple comparison among previous works, we also include the standard mAP metric, which has been widely used in both offline-TAL~\cite{actionformer, gtad} and online-TAL \cite{cagqil, oat} literature. We report mAPs with varying tIOUs in a set $\{0.3,0.4,0.5,0.6,0.7\}$ for THUMOS14 and $\{0.5,0.75,0.95\}$ for FineAction. An average value of those mAPs with multiple tIOUs is also reported for succinct comparison. For Epic-Kitchens 100 dataset, we evaluate mAP@$0.5$ using ``noun'' annotations, as they present a greater challenge compared to its verb annotations \cite{actionformer}. To evaluate the performance of ODAS, we measure the point-level average precision (p-AP) and calculate p-mAP by averaging p-AP over all action classes, following previous works~\cite{odas, startnet, cagqil}. Given that the ODAS task aims to achieve fast detection of action starts, we restrict the offset value to 3 seconds, as late detection beyond this threshold is considered not useful.

\subsection{Implementation Details}
\noindent\textbf{Architecture} 
%Our architecture is computationally efficient with the simple architecture as shown in Fig.~\ref{fig:main} (b).
Inspired by the latest research \cite{miniroad}, a stacked GRU \cite{gru} and MLPs with residual connection are used for the unidirectional sequential encoder and a state classifier respectively. 
This efficient recurrent design enables our model to process over \underline{500 fps}, underscoring its suitability for online applications.
Exploring effective architectural design choices is a promising direction, but we stick to the minimalist design in this paper as it is not our primary focus.
%While devising an efficient yet effective model architecture is an interesting research direction, it is not our primarly research focus in this paper so we will stick in to the aforementioned minimalist design choice.

\noindent\textbf{Class label} While the ActionSwitch framework focuses on class-agnostic action instances and our main F1 metric does not require class and confidence scores for evaluation, the class labels and corresponding confidence scores are necessary to measure per class mAP.
Hence, we train an extra classifier that takes feature sequences as its input and predicts the class labels for the given sequences. We take a vanilla transformer \cite{transformer} as its architecture, and the max class logit is directly treated as the confidence score. 
To ensure a fair comparison, we apply this same classifier to previous class-agnostic methods, resulting in slightly better performance than reported in the original paper.

\input{table/fineaction_tal}
\input{table/ek100}

\subsection{Main Results}
Tab. \ref{tab:fineaction_main}, Tab. \ref{tab:ek100} and Tab. \ref{tab:thumos_main} present the main experimental results of our ActionSwitch framework. For clarity, the performance of several state-of-the-art offline TAL methods is also reported. Here, the standard mAP metric is observed to have a stronger correlation with recall than precision (see supplementary materials for further discussion). Therefore in offline TAL methods, while using all action proposals without filtering boosts mAP scores, it becomes crucial to eliminate low-confidence proposals to achieve a reasonable F1 score.
\input{table/thumos14_tal}

While there is still a performance gap compared to the state-of-the-art offline TAL method \cite{actionformer}, ActionSwitch sets a new standard for both mAP and F1 scores among OAD-extended On-TAL methods. 
Unlike CAG-QIL \cite{cagqil} which requires two-stage training and SimOn \cite{simon} which depends on an On-TAL specific architecture, ActionSwitch offers a straightforward extension from the OAD framework to On-TAL with a minor modification of the output layer and incorporation of additional loss term. 
Notably as the number of classes grows, SimOn's class-aware grouping strategy significantly falters with the excessive number of proposals and low precision (Tab. \ref{tab:fineaction_main} and \ref{tab:ek100}), showing the limitations of classwise threshold-based grouping. 
In contrast, our method shows balanced precision and recall across all the datasets, and the use of the \texttt{argmax} operation streamlines the process by eliminating the need to predefine the threshold.

Furthermore, the TAL extension method, OAT~\cite{oat}, achieves impressive performance on relatively small datasets (\eg,~THUMOS14) but experiences training instability when applied to large-scale datasets that contain temporally varying action instances.
For example, in the FineAction dataset (Tab. \ref{tab:fineaction_main}), OAT~\cite{oat} performs reasonably well only when the temporal resolution is downsampled, and even in that case, our ActionSwitch performs better.

Also on Epic-Kitchens 100 (Tab. \ref{tab:ek100}), ActionSwitch surpasses OAT in both F1 and mAP scores, as OAT shows a stark imbalance between precision and recall.
This is significant as our ActionSwitch also has the inherent ability to handle ODAS tasks, a functionality not supported by OAT. (Section \ref{sec:probsetting})

\input{table/thumos14_oad}

We present the ODAS performance in Tab.~\ref{tab:odas}, which shows that our method achieves a new state-of-the-art performance. 
Given that early detection is crucial in ODAS, our method's superior performance at lower offsets highlights its strength in scenarios where immediate action detection is vital. Note that reported \textit{m}AP is much higher than those in the original papers~\cite{cagqil, simon} as we utilized Kinetics pre-trained features in all three models.

\subsection{Ablation Studies}
\label{sec:ablation_studies}
\noindent\textbf{Number of switches and conservativeness loss} 
Tab.~\ref{tab:abl_loss_weight} presents comprehensive ablation studies on our ActionSwitch setting. 
The 1-switch setting with $\alpha=0$ is considered the baseline for our approach which is a naive OAD grouping extension. 
By adopting one additional switch, the model can capture overlapping action instances. 
The recall of the 2-switch setting is much higher than its 1-switch counterpart, supporting the aforementioned claim. 
This tendency is even more apparent in the FineAction dataset, which contains more overlapping action instances than THUMOS14. 
\input{table/abl_loss_weight}
However, the precision remains low due to the grouping error that is prone in OAD extension On-TAL approaches.
Here, conservativeness loss significantly alleviates the grouping error.
We observe a tradeoff relationship between precision and recall when applying varying alpha values.
A higher $\alpha$ value indicates a stronger conservativeness prior, resulting in better precision but lower recall, and vice versa. 
An adequate $\alpha$ value leads to balanced precision and recall, leading to the best F1 value.

\noindent\textbf{Can we include 3 or more switches?} 
Expanding our framework to incorporate three or more action switches is straightforward; it only requires adding additional states corresponding to the combination of switches.
Yet, in most of the established TAL benchmark datasets \cite{THUMOS14, fineaction, epickitchen}, over 99\% of the frames have no more than two simultaneous actions, providing almost no ground truth data for scenarios where a third switch would be activated.

\input{table/multithumos_new}
\begin{figure}[b!]
    \centering
    \includegraphics[width=0.9\textwidth]{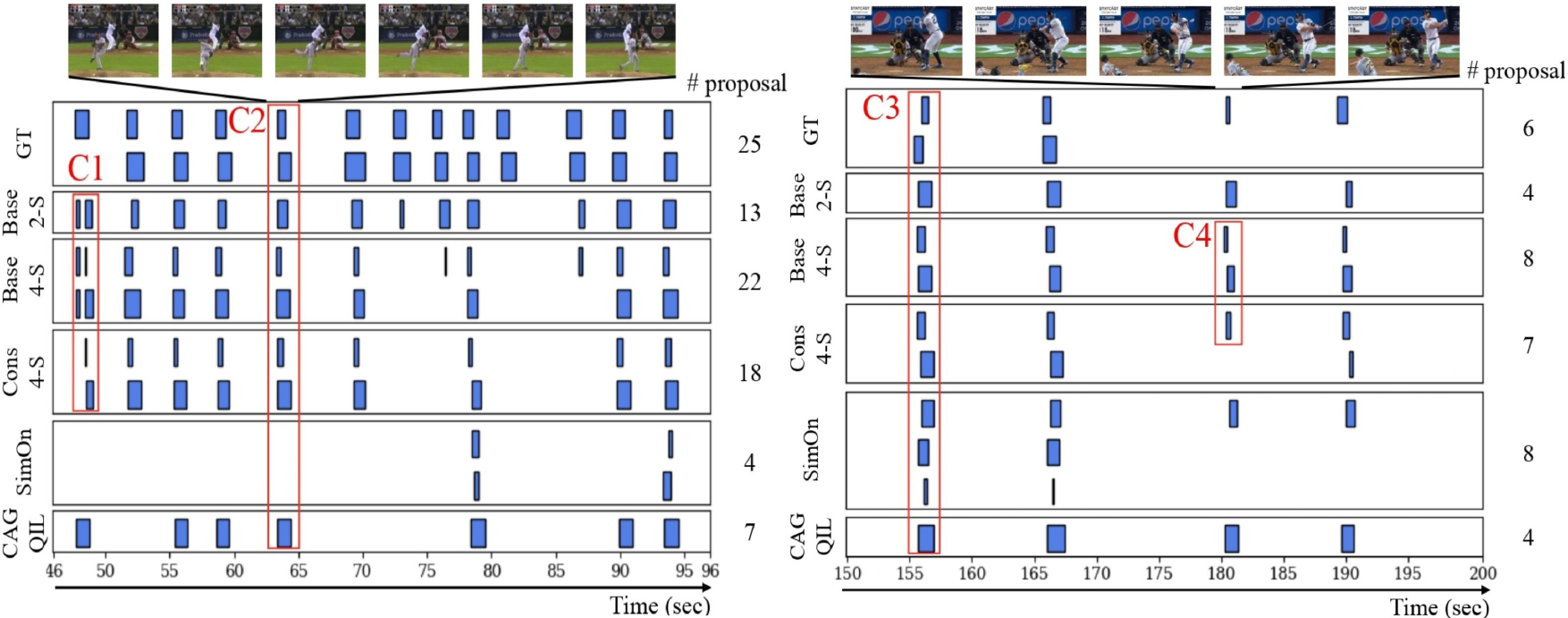}
    \caption{
    Qualitative results of On-TAL models. 
    If the action instances are overlapped, they are placed in other lines.
    We show ground-truth (GT) and the output of 2-state (2S), 4-state (4S), conservative loss (Cons), SimOn~\cite{simon} and CAG-QIL~\cite{cagqil}.
    Refer to Sec.~\ref{sec:qual_comp} for an analysis of four cases (\textcolor{red}{C1$\sim$C4}) which are annotated by the \textcolor{red}{red boxes}.
    }
    \label{fig:qual_comp}
    \label{fig:teaser}
\end{figure}
The Multithumos dataset \cite{multithumos}, however, is characterized by significant overlap among multiple action instances, despite its smaller scale. 
Therefore, we chose this dataset to perform an ablation study on configurations with three or more switches, with the results presented in Tab. \ref{tab:multithumos}.
The result not only shows that our method established a new state-of-the-art performance in the Multithumos dataset, but also highlights critical insights. 
Similar patterns to those in Tab. \ref{tab:abl_loss_weight} emerge here, but due to the dataset's dense overlaps, the optimal F1 score occurs with a \textbf{3 switch setup}. 
\textit{More switches enhance the model's ability to detect additional overlapping actions}, as indicated by the improved recall.
However, while introducing the conservativeness loss as an auxiliary term can mitigate the issues, it also leads to a precision trade-off. 
This becomes evident when the model overestimates action instances, as shown by comparing the number of predicted proposals with the actual ground truth counts.

\subsection{Qualitative Results}
\label{sec:qual_comp}
In Fig.~\ref{fig:qual_comp}, we demonstrate the effectiveness of our proposed components. Specifically, \textcolor{red}{C1} illustrates how the conservativeness loss effectively addresses the action fragmentation issue. Additionally, \textcolor{red}{C2} and \textcolor{red}{C3} demonstrate the ability of our ActionSwitch framework to detect overlapping instances. 
\textcolor{red}{C3} shows that SimOn~\cite{simon} can detect multiple overlapping actions, but its detection is restricted by the predefined action classes, thus preventing it from detecting overlapping instances of the same action class.
In \textcolor{red}{C4}, the conservativeness loss leads the model to keep predicting state 1 instead of 3, preventing the excessive proposal generation.

%% file: table/fineaction_tal.tex
\begin{table*}[b!]
\centering
\caption{Comparison to other TAL methods on FineAction dataset~\cite{fineaction}. We report \textit{m}AP at different tIoU thresholds and average \textit{m}AP in [0.5:0.05:0.95]. 
\textsuperscript{\textdagger} indicates the results with the optimal threshold that drops the low-confidence proposals and achieves the best F1 score.
For each metric, the best is bolded.
}
\resizebox{0.97\linewidth}{!}{
    \begin{tabular}{@{} c|c | l| cc|cccc@{}}
    \toprule[2pt]
    \multicolumn{2}{c|}{Setting} & Method & F1 & Recall & mAP@0.5 & 0.75 & 0.95 & Avg. \\
    \toprule[1pt]
    \multicolumn{2}{c|}{\multirow{2}{*}{\begin{tabular}{c} Offline\\TAL \end{tabular}}} & ActionFormer~\cite{actionformer} & 3.62 & \textbf{62.83} & \textbf{21.21} & \textbf{11.02} & \textbf{1.68} & \textbf{11.74} \\
    \multicolumn{2}{c|}{} & ActionFormer\textsuperscript{\textdagger}~\cite{actionformer} & \textbf{30.14} & 25.25 & 14.13 & 9.05 & 1.57 & 9.16 \\
    \midrule[1pt]
    \multirow{5}{*}{\begin{tabular}{c} Online\\TAL\end{tabular}} & \multirow{2}{*}{TAL-Extension} & OAT (\textit{downsampled by 4})~\cite{oat} & \textbf{18.64} & \textbf{38.10} & \textbf{8.88} & \textbf{1.56} & \textbf{0.02} & \textbf{3.04} \\
    & & OAT~\cite{oat} & 7.21 & 8.14 & 1.19 & 0.12 & 0.00 & 0.34 \\
    \cdashline{2-9}
    \noalign{\vskip 0.5ex}
    & \multirow{3}{*}{OAD-Extension} & CAG-QIL~\cite{cagqil} & 15.67 & 12.73 & 8.00 & 4.07 & \textbf{1.05} & 4.45 \\
    & & SimOn~\cite{simon} & 10.53 & 18.06 & 7.95 & 3.53 & 0.98 & 4.12 \\
    & & \textbf{ActionSwitch (Ours)} & \textbf{19.44} & \textbf{21.76} & \textbf{10.58} & \textbf{4.71} & 0.64 & \textbf{5.36} \\
    \bottomrule[2pt]
\end{tabular}
}
\label{tab:fineaction_main}
% \vspace{-3mm}
\end{table*}

%% file: table/ek100.tex
\begin{table}[t!]
\centering
\caption{Comparison on Epic-Kitchens 100 dataset \cite{epickitchen}. 
\# proposal denotes the number of generated proposals of each method, and \# ground truth refers to the number of ground truth proposals.
We report class-specific \textit{m}AP at $tiou=0.5$ using ``noun'' annotations. 
SimOn \cite{simon} and OAT \cite{oat} exploit class information for action instance generation, in contrast to OAD-Grouping \cite{cagqil}, CAG-QIL \cite{cagqil}, and our method, which do not.
For each metric, the best is bolded.}
\resizebox{0.97\linewidth}{!}{
\begin{tabular}{cccccccc}
\toprule[2pt]
% \multicolumn{8}{c}{EK100} \\ \toprule[1.5pt]
\multicolumn{1}{c|}{Setting} &
  \multicolumn{1}{c|}{Method} &
  \multicolumn{1}{c|}{F1} &
  \multicolumn{1}{c|}{precision} &
  \multicolumn{1}{c|}{recall} &
  \multicolumn{1}{c|}{mAP@0.5} &
  \multicolumn{1}{c|}{\# proposal} &
  \# ground truth \\ \midrule[1pt]
\multicolumn{1}{c|}{\begin{tabular}[c]{@{}c@{}}TAL-extension\end{tabular}} &
  \multicolumn{1}{l|}{OAT \cite{oat}} &
  \multicolumn{1}{c|}{27.583} &
  \multicolumn{1}{c|}{17.595} &
  \multicolumn{1}{c|}{\textbf{63.798}} &
  \multicolumn{1}{c|}{3.296} &
  \multicolumn{1}{c|}{35054} &
  \multirow{5}{*}{9668} \\ \cdashline{1-7}
\multicolumn{1}{c|}{\multirow{4}{*}{\begin{tabular}[c]{@{}c@{}}OAD-extension\end{tabular}}} &
  \multicolumn{1}{l|}{CAG-QIL \cite{cagqil}} &
  \multicolumn{1}{c|}{23.117} &
  \multicolumn{1}{c|}{21.347} &
  \multicolumn{1}{c|}{25.206} &
  \multicolumn{1}{c|}{2.442} &
  \multicolumn{1}{c|}{11416} &
   \\
\multicolumn{1}{c|}{} &
  \multicolumn{1}{l|}{SimOn \cite{simon}} &
  \multicolumn{1}{c|}{4.395} &
  \multicolumn{1}{c|}{2.351} &
  \multicolumn{1}{c|}{33.481} &
  \multicolumn{1}{c|}{1.846} &
  \multicolumn{1}{c|}{137685} &
   \\
\multicolumn{1}{c|}{} &
  \multicolumn{1}{l|}{OAD-Grouping \cite{cagqil}} &
  \multicolumn{1}{c|}{21.416} &
  \multicolumn{1}{c|}{25.533} &
  \multicolumn{1}{c|}{18.442} &
  \multicolumn{1}{c|}{2.267} &
  \multicolumn{1}{c|}{6983} &
   \\
\multicolumn{1}{c|}{} &
  \multicolumn{1}{l|}{\textbf{ActionSwitch (Ours)}} &
  \multicolumn{1}{c|}{\textbf{32.444}} &
  \multicolumn{1}{c|}{\textbf{29.858}} &
  \multicolumn{1}{c|}{35.519} &
  \multicolumn{1}{c|}{\textbf{3.597}} &
  \multicolumn{1}{c|}{11501} &
   \\ \bottomrule[2pt]
\end{tabular}%
}
\label{tab:ek100}
% \vspace{-5mm}
\end{table}

%% file: table/thumos14_tal.tex
\begin{table}[t!]
\centering
\caption{Comparison to other TAL methods on THUMOS14 dataset~\cite{THUMOS14}.
We report \textit{m}AP at different tIoU thresholds and average \textit{m}AP in [0.3:0.1:0.7]. 
In the offline TAL results, \textsuperscript{\textdagger} indicates the results with the optimal threshold that drops the low-confidence proposals and achieves the best F1 score.
OAT \cite{oat} performs On-TAL within a more flexible constraint than other works \cite{cagqil, simon}. (Section \ref{sec:probsetting})
For each metric, the best is bolded.
}
\resizebox{0.9\linewidth}{!}{
    \begin{tabular}{@{} c|c | l | cc | cccccc@{}}
    \toprule[2pt]
    \multicolumn{2}{c|}{Setting} & Method & F1 & Recall & mAP@0.3 & 0.4 & 0.5 & 0.6 & 0.7 & Avg. \\
    \toprule[1pt]

    \multicolumn{2}{c|}{\multirow{6}{*}{\begin{tabular}{c} Offline\\TAL\end{tabular}}} & G-TAD~\cite{gtad} & 6.4 & 83.4 & 58.8 & 52.2 & 43.6 & 33.3 & 22.9 & 42.2 \\ 
    \multicolumn{2}{c|}{} & TadTR~\cite{tadtr} & 3.5 & 88.1 & 68.8 & 62.7 & 55.9 & 45.3 & 32.3 & 53.0 \\  
    \multicolumn{2}{c|}{} & ActionFormer~\cite{actionformer} & 13.8 & \textbf{94.0} & \textbf{77.5} & \textbf{73.5} & \textbf{66.0} & \textbf{55.4} & \textbf{40.6} & \textbf{62.6} \\ % TSN feature
    \multicolumn{2}{c|}{} & G-TAD\textsuperscript{\textdagger}~\cite{gtad} & 51.1 & 49.9 & 45.6 & 40.7 & 34.4 & 26.7 & 18.7 & 33.2 \\ 
    \multicolumn{2}{c|}{} & TadTR\textsuperscript{\textdagger}~\cite{tadtr} & 68.4 & 63.0 & 55.4 & 51.4 & 46.5 & 38.4 & 28.0 & 43.9 \\ 
    \multicolumn{2}{c|}{} & ActionFormer\textsuperscript{\textdagger}~\cite{actionformer} & \textbf{75.5} & 73.6 & 66.9 & 63.2 & 56.3 & 47.4 & 35.1 & 53.8 \\ % TSN feature
    
    % \multirow{6}{*}{\begin{tabular}{c} Offline\\TAL\end{tabular}} & & G-TAD~\cite{gtad} & 6.4 & 83.4 & 58.8 & 52.2 & 43.6 & 33.3 & 22.9 & 42.2 \\ 
    % & & TadTR~\cite{tadtr} & 3.5 & 88.1 & 68.8 & 62.7 & 55.9 & 45.3 & 32.3 & 53.0 \\  
    % & & ActionFormer~\cite{actionformer} & 13.8 & 94.0 & 77.5 & 73.5 & 66.0 & 55.4 & 40.6 & 62.6 \\ % TSN feature
    % & & G-TAD\textsuperscript{\textdagger}~\cite{gtad} & 51.1 & 49.9 & 45.6 & 40.7 & 34.4 & 26.7 & 18.7 & 33.2 \\ 
    % & & TadTR\textsuperscript{\textdagger}~\cite{tadtr} & 68.4 & 63.0 & 55.4 & 51.4 & 46.5 & 38.4 & 28.0 & 43.9 \\ 
    % & & ActionFormer\textsuperscript{\textdagger}~\cite{actionformer} & 75.5 & 73.6 & 66.9 & 63.2 & 56.3 & 47.4 & 35.1 & 53.8 \\ % TSN feature
    
    \midrule[1pt]
    
    \multirow{4}{*}{\begin{tabular}{c} Online\\TAL\end{tabular}} & TAL-Extension & OAT~\cite{oat} & \textbf{62.9} & \textbf{70.3} & \textbf{64.1} & \textbf{57.4} & \textbf{47.8} & \textbf{36.7} & \textbf{20.3} & \textbf{45.3} \\
    \cdashline{2-11}
    \noalign{\vskip 0.5ex}
    & \multirow{3}{*}{OAD-Extension} & CAG-QIL~\cite{cagqil} & 45.8 & 50.4 & 48.8 & 40.9 & 33.6 & 24.9 & 17.3 & 33.1 \\
    % & & SimOn~\cite{simon} & 24.8 & 49.6 & 48.7 & 39.7 & 29.8 & 21.5 & 13.2 & 30.6 \\ % our feature 
    & & SimOn~\cite{simon} & 28.7 & 52.0 & 52.2 & 43.6 & 32.3 & 22.5 & 14.2 & 33.0 \\ % simon feature
    & & \textbf{ActionSwitch (Ours)} & \textbf{53.2} & \textbf{60.1} & \textbf{57.2} & \textbf{50.8} & \textbf{41.7} & \textbf{30.5} & \textbf{21.3} & \textbf{40.3} \\
    \bottomrule[2pt]
\end{tabular}
}
\label{tab:thumos_main}
\end{table}

%% file: table/thumos14_oad.tex
\begin{wraptable}{r}{0.51\textwidth}
\begin{minipage}{0.51\textwidth}
\centering
\caption{Comparison of other SOTA ODAS methods on THUMOS14 dataset~\cite{THUMOS14}. 
}
\resizebox{0.8\linewidth}{!}{
\begin{tabular}{@{}l |ccc @{}}
\toprule[2pt]
\multirow{2}{*}{ODAS Method} & \multicolumn{3}{c}{Offsets} \\ \cline{2-4} 
 & 1 & 2 & 3 \\
\midrule[1pt]
CAG-QIL~\cite{cagqil} & 28.30 & 42.86 & 50.12 \\
SimOn~\cite{simon} & 31.45 & 46.22 & 54.11 \\
\textbf{ActionSwitch (Ours)} & \textbf{33.06} & \textbf{47.06} & \textbf{54.44} \\
\bottomrule[2pt]
\end{tabular}%
}
\label{tab:odas}
\end{minipage}
\end{wraptable}

%% file: table/abl_loss_weight.tex
\begin{table}[t!]
\centering
\caption{Ablation study of the number of states in ActionSwitch and the conservativeness loss whose weight is $\alpha$ as shown in Eq.~\ref{eq:loss_function}.
}
\resizebox{0.8\textwidth}{!}{
\begin{tabular}{@{} c| c |cccc | c |cccc @{}}
\toprule[2pt]
\multirow{2}{*}{\# switch} & \multicolumn{5}{c|}{Thumos14~\cite{THUMOS14}} & \multicolumn{5}{c}{FineAction~\cite{fineaction}} \\
\cmidrule[1pt]{2-6} \cmidrule[1pt]{7-11}
& $\alpha$ & F1 & Recall & Precision & \# proposal &  $\alpha$ & F1 & Recall & Precision & \# proposal \\
\midrule[1pt]
\multirow{4}{*}{1} & 0.000 & 42.67 & \textbf{59.49} & 33.26 & 6006
& 0.000 & 14.53 & \textbf{14.71} & 14.35 & 24846 \\
& 0.010 & 49.25 & 57.77 & 42.92 & 4519
& 0.005 & 15.78 & 14.45 & 17.37 & 20159 \\
& \textbf{0.025}  & \textbf{50.41} & 56.04 & 45.81 & 4108
& \textbf{0.010}  & \textbf{16.53} & 13.28 & \textbf{21.90} & 14694 \\
& 0.050 & 49.91 & 49.61 & \textbf{50.21} & 3318
& 0.025 & 9.34 & 6.28 & 18.17 & 8380 \\ 
\midrule[1pt]
\multirow{4}{*}{2} & 0.000 & 45.87 & \textbf{63.96} & 35.75 & 6007 & 0.000 & 17.57 & \textbf{22.36} & 14.46 & 37469 \\
& 0.010 & 49.28 & 61.25 & 41.23 & 4989
& 0.005 & 18.32 & 22.00 & 15.70 & 33949 \\
& \textbf{0.025} & \textbf{53.20} & 60.10 & 47.73 & 4228
& \textbf{0.010} & \textbf{19.44} & 21.76 & 17.56 & 29972 \\
& 0.050 & 50.20 & 52.76 & \textbf{47.88} & 3701
& 0.025 & 18.71 & 16.76 & \textbf{21.15} & 19205 \\
\bottomrule[2pt]
\end{tabular}
}
\label{tab:abl_loss_weight}
\end{table}

%% file: table/multithumos_new.tex
\begin{table}[t!]
%\vspace{-5mm}
\centering
\caption{Multiple switch configurations tested in Multithumos \cite{multithumos} dataset. Ablation study on conservativeness loss term $\mathcal{L}_c$ ($\alpha=0.025$) is also conducted in multi-switch settings. A good F1 score is observed when the number of generated predictions (\# proposal) is close to the number of ground truth instances (\# GT\_proposal).}
\resizebox{0.8\linewidth}{!}{
\begin{tabular}{c|cc|cccc|c}
\toprule[2pt]
Method & \# switch & $\mathcal{L}_c$ & F1 & Precision & Recall & \# proposal & \# GT\_proposal \\
\midrule[1pt]
\multirow{6}{*}{\textbf{ActionSwitch (Ours)}} & 1 & o & 16.51 & 42.85 & 10.22 & 4816 & \multirow{11}{*}{20186} \\
 & 2 & o & 30.82 & 34.37 & 27.94 & 16405 &  \\
 &\cellcolor{gray!25}3 &\cellcolor{gray!25}o & \cellcolor{gray!25}\textbf{32.76} &\cellcolor{gray!25}30.04 &\cellcolor{gray!25}36.02 &\cellcolor{gray!25}24211 &  \\
 & 4 & o & 29.25 & 22.67 & 41.21 & 36682 &  \\
 \cdashline{2-7}
 & 3 & x & 26.63 & 20.53 & 37.91 & 37287 &  \\
 & 4 & x & 23.16 & 16.02 & 41.86 & 52758 &  \\
\cmidrule[1pt]{1-7}
CAG-QIL~\cite{cagqil} & \multicolumn{2}{c|}{-} & 22.10 & \textbf{48.7} & 14.29 & 5926 \\
\cmidrule[1pt]{1-7}
SimOn~\cite{simon} & \multicolumn{2}{c|}{-} & 20.59 & 12.42 & 60.17 & 97759 \\
\cmidrule[1pt]{1-7}
OAT~\cite{oat} & \multicolumn{2}{c|}{-} & 29.63 & 18.48 & \textbf{74.74} & 81626 \\
\bottomrule[2pt]
\end{tabular}
}
\label{tab:multithumos}
\end{table}

%% file: sec/5_conclusion.tex
\section{Discussion and Conclusion}
\label{sec:discussion}
\noindent\textbf{Towards an Open-Vocabulary Framework}
Existing On-TAL approaches are evaluated in a closed-vocabulary setting with predefined action classes. However, there is a growing interest in open-vocabulary research~\cite{vild, detic, cora} as real-world scenarios need deeper comprehension beyond the identification of pre-defined actions and events in datasets.
Class-aware On-TAL methods \cite{oat, simon} struggle in this context due to their reliance on preset classes.
ActionSwitch, on the other hand, presents a versatile, class-agnostic framework that facilitates integration with video-language models \cite{vificlip}. 
It paves the path for expansive vocabulary classification, a promising direction for future research.

\noindent\textbf{Is different instantiation possible?}
Instead of the state-emitting OAD approach, one can treat the output of the OAD model as a state transition, where 0 refers to maintaining the current state, while 1 and 2 indicate state transitions made by manipulating switch 1 and 2, respectively.
While this formulation can reduce the required state number, it brings up severe class imbalance due to the sparsity of action boundaries. (Sec.~\ref{sec:conservative_loss}) It will be promising future research to develop reinforcement learning algorithms~\cite{rlbook} with this formulation.

\section{Conclusion}
We introduce ActionSwitch, the first On-TAL method capable of detecting overlapping action instances without relying on class information. 
Coupled with novel conservativeness loss, experiments on multiple datasets demonstrate its effectiveness despite its simplicity.
We hope that our proposed ActionSwitch can serve as a strong baseline and inspire further On-TAL research.
\newline